\documentclass[pmlr]{jmlr}

\usepackage{longtable}

\usepackage{booktabs}
\usepackage[load-configurations=version-1]{siunitx} 

\theorembodyfont{\upshape}
\theoremheaderfont{\scshape}
\theorempostheader{:}
\theoremsep{\newline}

\jmlrvolume{1}
\jmlryear{2024}
\jmlrworkshop{AI for Education: Bridging Innovation and Responsibility}

\title[NoteAid EHR Interaction]{EHR Interaction Between Patients and AI\titlebreak NoteAid EHR Interaction} 

\usepackage{natbib}
\usepackage{amsmath}
\usepackage{amsfonts}
\usepackage{multirow}
\usepackage{tcolorbox}
\usepackage{booktabs}
\usepackage{graphicx}
\usepackage{caption}
\usepackage{longtable}

\author{\Name{Xiaocheng Zhang}\thanks{Indicates equal contribution}\Email{xiaochengzha@umass.edu}\\
\Name{Zonghai Yao}\footnotemark[1] \Email{zonghaiyao@umass.edu }\\
\Name{Hong Yu} \Email{Hong\_Yu@uml.edu }\\
\addr College of Information and Computer Sciences, University of Massachusetts Amherst}

\begin{document}

\maketitle

\begin{abstract}
With the rapid advancement of Large Language Models (LLMs) and their outstanding performance in semantic and contextual comprehension, the potential of LLMs in specialized domains warrants exploration. This paper introduces the NoteAid EHR Interaction Pipeline, an innovative approach developed using generative LLMs to assist in patient education, a task stemming from the need to aid patients in understanding Electronic Health Records (EHRs). Building upon the NoteAid work, we designed two novel tasks from the patient's perspective: providing explanations for EHR content that patients may not understand and answering questions posed by patients after reading their EHRs. We extracted datasets containing 10,000 instances from MIMIC Discharge Summaries and 876 instances from the MADE medical notes collection, respectively, executing the two tasks through the NoteAid EHR Interaction Pipeline with these data. Performance data of LLMs on these tasks were collected and constructed as the corresponding NoteAid EHR Interaction Dataset.
Through a comprehensive evaluation of the entire dataset using LLM assessment and a rigorous manual evaluation of 64 instances, we showcase the potential of LLMs in patient education. Besides, the results provide valuable data support for future exploration and applications in this domain while also supplying high-quality synthetic datasets for in-house system training~\footnote{To appear in AAAI2024 Workshop on AI for Education (AI4ED)}.
\end{abstract}

\section{Introduction}
\label{sec:intro}
The progress in healthcare extends beyond medical breakthroughs; it represents an advancement in patients' willingness to engage in self-care. While new medications and treatment modalities are undeniably effective, they necessitate active patient involvement and cooperation. Many diseases often require patients and their families to possess a clear understanding of caregiving \citep{HomeHealthCare, nurseNavigators}. The question of how to empower patients with a comprehensive understanding of their medical conditions and ensure their adherence to medical recommendations is a matter of considerable importance. In other words, patient education is of paramount importance. Currently, enabling patients to access Electronic Health Record (EHR) notes is an economically effective means of enhancing patient education. Initiatives such as ``The Patient-Centered Access to Secure Systems Online'' (PCASSO) \citep{GivingPatientsAccess} and ``NoteAid'' \citep{NoteAid} exemplify these efforts. This not only fosters patient empowerment but also strengthens the patient-physician partnership, ultimately leading to improved healthcare outcomes. 
\begin{figure}[h]
    \centering
    \floatconts
    {fig:task_example_introduction_pipeline_work_flow}
    {
    \caption{\textbf{Task Examples in NoteAid EHR Interaction Pipeline.}\\
    The figure (a) presented here illustrates the dialogues for both the Explanation Task and the Q\&A Task. The bolded text within the EHR-Discharge Instruction represents the points that the Mock Patient Agent selected for questioning and identified as challenging to comprehend. The conversation here revolves around the concept and application of the ``incentive spirometer.''
    The figure (b) represents our pipeline structure.}}
    {
        \subfigure[Task Example][c]{\label{fig:task_example_introduction}
        \includegraphics[width=0.41\textwidth, trim=3cm 3cm 4cm 1cm, clip]{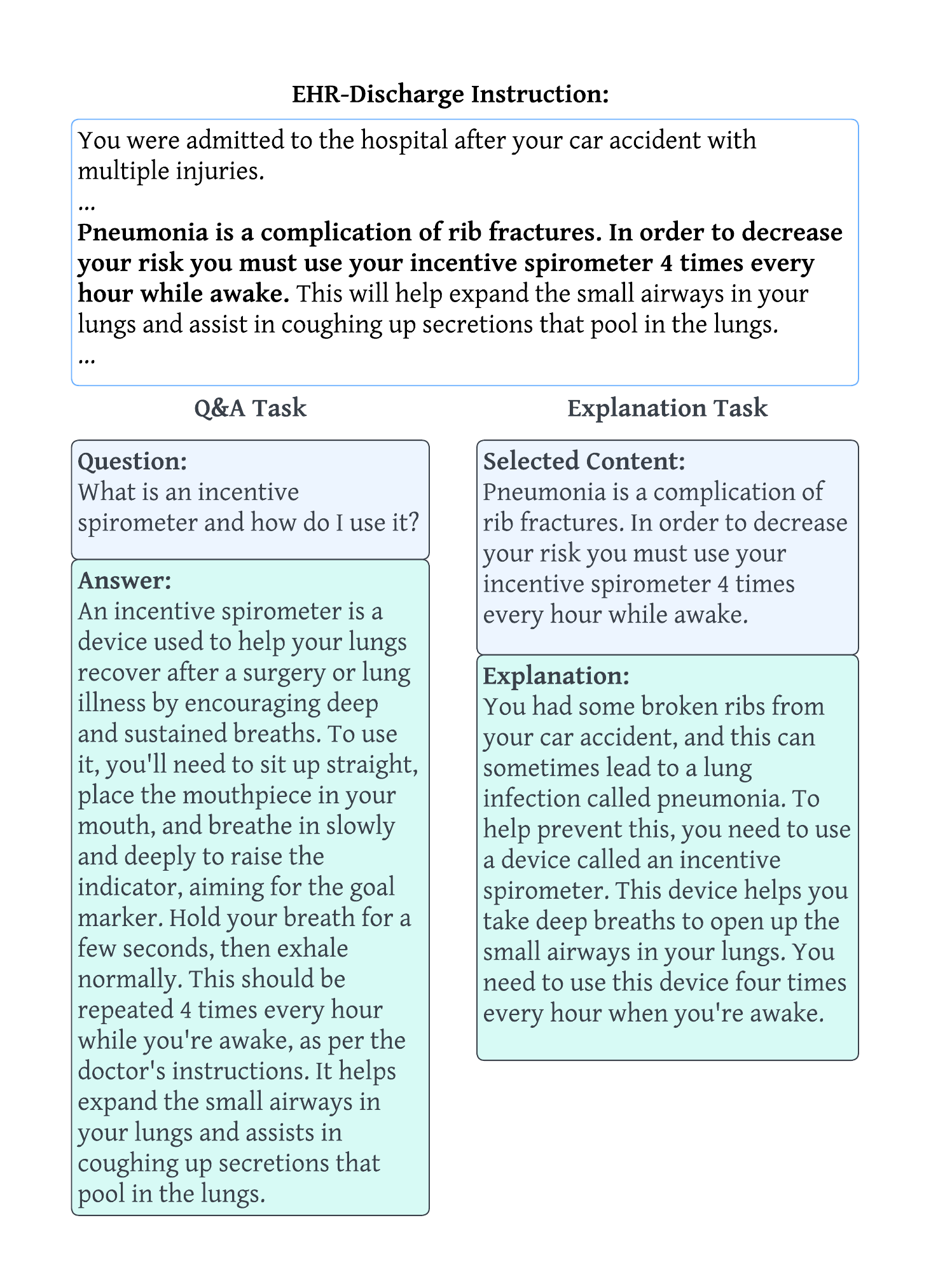}}
        \quad
        \subfigure[Pipeline Structure][c]{\label{fig:pipeline_work_flow}
        \includegraphics[width=0.5\textwidth, trim=0.7cm 0cm 0.8cm 0cm, clip]{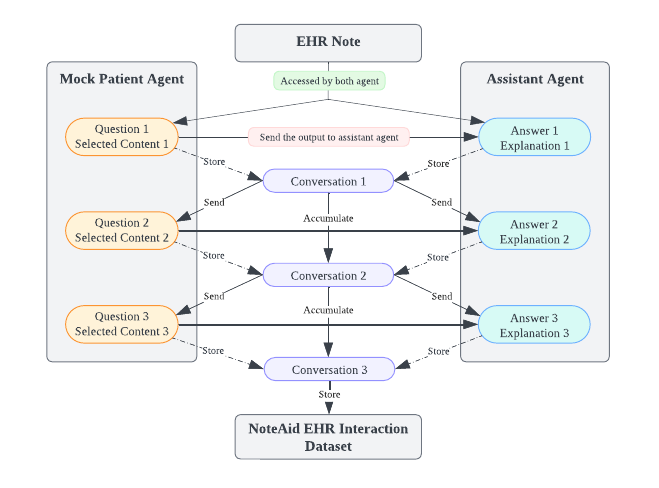}}
    }
    \vspace{-5mm}
\end{figure}

So far, many electronic medical note management projects have primarily focused on providing patients with secure and private access to their medical notes. Researchers like NoteAid have also incorporated Natural Language Processing (NLP) technology to predict and provide lay definitions for complex medical jargon within the notes \citep{NoteAidmedjex,NoteAidreadme}, effectively assisting patients in understanding the content of their medical notes \citep{NoteAid, NoteAidEffectiveness}. However, there is still untapped potential in these patient-oriented services offered by researchers like NoteAid. The medical jargon predictions are generated by language models directly rather than being selected by the patients themselves. There still exists a lack of subjective interaction between patients and their medical notes, and increasing such interaction can lead to a deeper understanding and engagement \citep{Whitehurst2002dialogic,cai2023paniniqa}. 

With the rapid development of LLMs, their integration with the field of medicine is poised to become a significant trend, as discussed in prior works \citep{brown2020language,openai2023gpt4}. 
So, our primary objective is to confirm whether LLMs can provide interactive services to patients regarding their medical notes and elevate the level of medical education for the patients. To achieve this, we introduce tasks related to EHR question-answering (Q\&A) and text explanation. Q\&A based on EHR notes has been demonstrated as an effective interaction method for enhancing patients' comprehension of medical instructions \citep{cai2023paniniqa,zhang2023ehrtutor}. The explanation of textual content serves as both a derivative action of Q\&A and a method to enhance memory retention. This activity aims to enhance patient engagement with their EHRs, providing them with additional avenues for interaction. We aim to encourage patients to proactively seek an understanding of treatment plans, including their details and rationale, while reading medical notes. When patients spontaneously express a desire to comprehend EHR notes, these two tasks can effectively fulfill the majority of their needs, thereby enhancing patients' subjective initiative in their healthcare process and improving their medical education.

In this study, we constructed a system named the NoteAid EHR Interaction Pipeline, harnessing the expertise and reasoning capabilities of LLMs \footnote{We used ChatGPT and GPT-4 in our experiments.} in the medical domain \citep{LLMReasoninginMedical, LLMDR, yang2023performance}. This pipeline employs an LLM as an agent, simulating the role of a medical assistant by engaging in conversation with users to perform EHR Q\&A and text explanation tasks. Inspired by recent advances in deploying two LLMs as cooperative agents for multi-round conversation generation \citep{Panait2005CooperativeML,Li2023CAMELCA,wang2023notechat}, we introduced another LLM model as an agent, portraying the role of a patient, to validate the effectiveness of LLMs in enhancing patients' medical education. This approach allows us to simulate plenty of instances where patients might interact with EHR notes, enabling the collection of data to create a synthetic NoteAid EHR Interaction Dataset. This dataset is utilized to assess the quality of LLM interactions and validate LLMs' ability to enhance patient education and train distillation models. Within this pipeline, both agents share the content of EHR notes and concurrently pass the conversation context to both agents, enabling them to maintain consistent contextual memory during the tasks. This mechanism enhances the quality of inquiries and responses.









\section{Related Work}
\subsection{Patient Education}
Patient education plays a crucial role in the success of therapy \citep{bastable2016essentials, golper2001patient}. Elevating the level of patient education has consistently been an important component within the healthcare system \citep{mccarthy2013understanding}. 
By providing clear and easily understandable medical information, patient education not only enhances patients' awareness of their own health conditions but also encourages them to actively participate in medical decision-making and self-management\citep{GRUMAN2010350, coulter2012patient}. From the perspective of delivering accurate and concise medical knowledge, we have designed Q\&A and text explanation tasks to enhance patients' levels of medical education.

\subsection{NoteAid}

NoteAid is a web-based natural language processing system designed to connect medical terminology in electronic health record (EHR) notes with simplified explanations, specifically tailored for easy comprehension by non-experts~\citep{NoteAid}. The NoteAid system comprises two core components: CoDeMed, a vocabulary resource providing layman's definitions for medical terms, and MedLink, a computational unit that associates medical terms with layman's definitions. 
In subsequent optimizations, the MedJEx~\citep{NoteAidmedjex} and README~\citep{NoteAidreadme} have been incorporated into the pipeline as a novel solution to predict medical jargon and generate lay definitions to streamline patient comprehension costs. 
In this paper, we introduce new interactive functionalities for NoteAid, enhancing patients' cognitive capabilities by utilizing LLMs to answer questions and provide explanations for medical content.

\subsection{LLMs for clinical synthetic data generation}
Recent research has yielded promising findings on using Large Language Models (LLMs) for data augmentation in tasks like code summarization, translation, and generation \citep{li2023feasibility, Dai2023ChatAugLC, zhou2022large, dai2022promptagator, yoo2021gpt3mix}. 
\citet{gilardi2023chatgpt} and \citet{ding2022gpt} investigated LLM efficacy in data annotation, demonstrating promising results on par with or surpassing human accuracy.
\citet{bonifacio2022inpars} used LLMs to generate training pairs for downstream models.
Within biomedicine, synthetic data generation is an active field in the clinical domain, especially to overcome privacy concerns or reduce the need for expert annotation~\citep{pereira2022secure,shafquat2022source,tang2023does, wang2023notechat, Liao2023DifferentiateCA,mishra2023synthetic,tran2023bioinstruct}.
The key findings are that LLMs demonstrate promising capabilities for data augmentation and annotation across domains. 
Their accuracy approximates or exceeds human performance in some tasks. Further research is needed to optimize their application and address concerns like privacy.

\subsection{LLMs for medical questions reasoning}
The experiments and researches prior \citep{liévin2023large, nori2023capabilities} has shown the reasoning ability of LLMs, such as GPT-3.5 and LLamma-2, in several medical datasets. The result indicates that LLMs could comprehend mostly complex medical questions. The application of using LLMs in clinical fields has potential \citep{singhal2023large, cascella2023evaluating}.

\section{Methods}

\subsection{Task Description}
Both the Q\&A task and the text explanation task (referred to as the Explanation task in the subsequent text) can be defined using the following approach: Assuming the given EHR Note $N: n_1, n_2,..., n_i$ represent $ith$ patient's note. Then we have patient's request $X: x_1^i, x_2^i,..., x_k^i$ where total $k$ requests will be inquired. Note that requests $X$ could be only a series of questions or a series of challenge EHR content. The response $Y: y_1^i, y_2^i, ..., y_k^i$ will be generated by the LLM $M_{large}$ with the corresponding prompt $P: p_1^i, p_2^i, ..., p_k^i$. We write $y_k^i = M_{large}(p_k^i)$ to indicate the process of response generated by the LLM with a prompt. Then we define a prompt generation function set $\mathcal{F}$ that generates $p$ in the following formula:\\
\begin{equation}P
  \begin{cases}
    p_1^i = f_{init}(n_i, x_1^i) \text{ for } f_{init} \in \mathcal{F}\\
    p_k^i = f(p_{k-1}^i, y_{k-1}^i, x_k^i) \text{ for } k \geq 1, \text{ and } f \in \mathcal{F}
  \end{cases}
\end{equation}\\
The prompt generation function basically adds up the word tokens $N$, $X$, and $P$ with several fixed prompt tokens in the way $f_{init}(n_i, x_1^i) = t_1 + n_i + t_2 + x_1^i$, $f(p_{k-1}^i, y_{k-1}^i, x_k^i) = p_{k-1}^i + t_3 + y_{k-1}^i + t_4 + x_k^i$, via $t \in T$ are fixed prompt tokens which can be found in appendix with the name called Generation Prompt. $Y$ is also required when $k > 1$ since the conversation includes the context above. So we claim that our NoteAid EHR Interaction Pipeline generates the response $Y$ as the following formula:\\
\begin{equation}Y
  \begin{cases}
    y_1^i = M_{large}(f_{init}(n_i, x_1^i)) \text{ for } f_{init} \in \mathcal{F}\\
    y_k^i = M_{large}(f(p_{k-1}^i, y_{k-1}^i, x_k^i)) \text{ for } k \geq 1, \text{ and } f \in \mathcal{F}
  \end{cases}
\end{equation}
\subsection{Pipeline Description} 
Sent the EHR to both agents in the pipeline simultaneously. The Mock Patient Agent generates questions in Q\&A task or selects challenging text in Explanation task. Subsequently, the Assistant Agent determines the response content based on the EHR and the output from the Mock Patient Agent. As illustrated in \figureref{fig:pipeline_work_flow}, the first round of conversation includes the EHR Note, along with the inputs and outputs of the two agents. In the second round of conversation, the historical records of each agent from the first round are differentiated, organized, and utilized as references for the same operations. This involves prompting the Mock Patient Agent to initiate the second round task. This structured conversation proceeds through three rounds, with its content, after removing redundant context, being stored as one instance within the NoteAid EHR Interaction Dataset.
\section{Experiment}
\subsection{Dataset Source}
The data used in our study is sourced from two primary datasets: a subset of discharge instructions from MIMIC-III\footnote{We used Microsft Azure GPT in this research.} \citep{Mimic-III}, totaling 10,000 records has been used, and a subset from MADE 1.0 \citep{MADE}, totaling 876 records. MIMIC-III is a publicly available medical information database encompassing a wide range of data pertaining to intensive care unit (ICU) patients. MADE, which stands for Medication, Indication, and Adverse Drug Events 1.0 corpus, is sourced from the National Center for Biotechnology Information (NCBI) disease corpus \citep{NLPMADE}.

\subsection{NoteAid EHR Interaction}
The discharge instructions from MIMIC-III, consisting of 10,000 records, was organized as follows: 8,000 records were allocated for the train data, 1,000 records for the validation data, and another 1,000 records for the test data. The remaining 876 records were sourced from MADE and were not subjected to the aforementioned allocation. Both the Q\&A and Explanation tasks were executed on the datasets descried above using the NoteAid EHR Interaction Pipeline (NIP) with GPT-3.5-Turbo (referred to as Turbo NIP) and GPT-4 (referred to as GPT-4 NIP). Therefore, each task comprised a total of 21,752 instances, with each instance containing three rounds of dialogue, involving either Q\&A or Explanation. In summary, the entire NoteAid EHR Interaction Dataset encompasses 43,504 instances, with Turbo NIP executing 10,876 Q\&A instances and 10,876 explanation instances. The quantity of instances executed by GPT-4 NIP matches that of Turbo NIP. For more dataset details, please refer to \tableref{tab:datasetStatistic}.


\setlength{\abovecaptionskip}{0pt}

\begin{table}[htbp]
\centering
\floatconts
{tab:datasetStatistic}
{\caption{\textbf{NoteAid EHR Interaction Statistic Table}\\This table presents fundamental statistical data for the NoteAid EHR Interaction. The ``tokens length'' is derived using the encoding algorithm of the GPT-3.5-Turbo model. ``Patient Agent'' and ``Assistant Agent'' represent the data collected by the respective agents in the NIP. For example, ``14.64 (14)'' indicates that during the execution of the Q\&A task by Turbo NIP on MIMIC-III, the Mock Patient Agent had an average tokens length of 14.64 with a median of 14.}}
{\resizebox{0.6\textwidth}{!}{\begin{tabular}{p{2cm} @{}clcccc}
                           &                    & \multicolumn{2}{c}{MIMIC-III}                   & \multicolumn{2}{c}{MADE}                        \\ \hline
                           &                    & Q\&A                   & Explanation            & Q\&A                    & Explanation           \\ \hline
Turbo NIP                  &                    & 10876                  & 10876                  & 10876                   & 10876                 \\
GPT-4 NIP                  &                    & 10876                  & 10876                  & 10876                   & 10876                 \\ \cline{3-6} 
\multicolumn{1}{l}{}       &                    & \multicolumn{2}{c}{avg. tokens length (median)} & \multicolumn{2}{c}{avg. tokens length (median)} \\ \hline
\multicolumn{1}{l}{}       &                    & Q\&A                   & Explanation            & Q\&A                    & Explanation           \\ \hline
\multirow{2}{*}{Turbo NIP} & Patient Agent & 14.64 (14)             & 22.84 (18)             & 15.94 (15)              & 27.92 (23)            \\
                           & Assistant Agent    & 63.36 (61)             & 85.19 (79)             & 93.41 (87)              & 99.61 (96)            \\
\multirow{2}{*}{GPT-4 NIP} & Patient Agent & 18.62 (18)             & 22.22 (19)             & 18.28 (17)              & 26.42 (24)            \\
                           & Assistant Agent    & 81.98 (78)             & 51.3 (47)              & 105.78 (101)            & 74.72 (69)           
\end{tabular}}}
\vspace{- 8mm}
\end{table}

\subsection{Evaluation and Results}
To evaluate the quality of our pipeline in performing the Q\&A and the Explanation task, we conducted both LLM evaluation and human evaluation. The criteria \tableref{tab:criteria} and quality level in \figureref{fig:qualityLevel} for evaluation were established after discussions with medical students who have clinical experience. During the evaluation process, due to resource constraints, we initially prioritized LLM evaluation. Subsequently, we performed human evaluation with a focus on dataset identified based on the assessment results from LLM evaluation.

\paragraph{LLM Evaluation} 
We utilized the LLMs' ability in medical questions reasoning to evaluate the performance of collected conversation. We randomly selected 100 instances from the 876 generated based on MADE and combined them with 1000 test and 1000 validation data generated from MIMIC-III for LLM evaluation (totally 2100 cases). The LLMs used for evaluation were GPT-3.5-Turbo (referred to as Turbo in this section) and GPT-4. We aggregated the criteria into an evaluation prompt \tableref{tab:generationPrompt} to guide the LLMs in scoring each conversation. The results are shown in the \tableref{tab:evalOverview}. From the figure, it is evident that Turbo received higher scores than GPT-4, particularly in the Turbo NIP Explanation task. This discrepancy can be attributed to the strict evaluation criteria outlined in our evaluation prompt. Turbo, while slightly less capable than GPT-4, did not fully meet these criteria. Additionally, we observed that data generated by GPT-4 NIP, whether in the Q\&A task or the Explanation task, exhibited more stable scores across both Turbo and GPT-4 evaluations. Notably, GPT-4 received a noticeably lower score in the Turbo NIP Explanation task when compared to Turbo.

        

\begin{table}[h]
\centering
\floatconts
{tab:evalOverview}
{\caption{\textbf{LLM Evaluation Overview} and \textbf{Human Evaluation Overview}\\Given an example of human evaluation, the distribution of data in the Q\&A task, sorted by quality level scores from high to low, is as follows: 95.96\%, 1\%, 1\%, 0\%, 2\%, 0\%. For the explanation task, the distribution by quality level is: 80.8\%, 12.12\%, 4.04\%, 1\%, 1\%, 1\%. }}
{\resizebox{0.8\textwidth}{!}{
\begin{tabular}{cclcccccc}
Evaluation Overview          &             & \multicolumn{1}{c}{Quality Level (\%)} & 5     & 4     & 3     & 2     & 1    & 0    \\ \hline
GPT-3.5-Turbo eval Turbo NIP & Q\&A        &                                        & 92.18 & 7.36  & 0.061 & 0.091 & 0.15 & 0.15 \\
                             & Explanation &                                        & 92.64 & 7.21  & 0.091 & 0.061 & 0    & 0    \\
GPT-3.5-Turbo eval GPT-4 NIP & Q\&A        &                                        & 90.21 & 9.52  & 0.091 & 0.15  & 0.03 & 0    \\
                             & Explanation &                                        & 94.7  & 5.18  & 0.12  & 0     & 0    & 0    \\
GPT-4 eval Turbo NIP         & Q\&A        &                                        & 88.73 & 9.21  & 0.03  & 1.15  & 0.55 & 0.33 \\
                             & Explanation &                                        & 57.5  & 40.12 & 1.85  & 0.18  & 0.03 & 0.3  \\
GPT-4 eval GPT-4 NIP         & Q\&A        &                                        & 92.48 & 7.48  & 0     & 0.03  & 0    & 0    \\
                             & Explanation &                                        & 89    & 10.94 & 0.06  & 0     & 0    & 0    \\
Human eval GPT-4 NIP         & Q\&A        &                                        & 95.96 & 1.01  & 1.01  & 0     & 2.02 & 0    \\
                             & Explanation &                                        & 80.81 & 12.12 & 4.04  & 1.01  & 1.01 & 1.01
\end{tabular}}}
\vspace{-6mm}
\end{table}

\paragraph{Human Evaluation} 
Because our LLM evaluation has shown that the data generated by GPT-4 NIP is of higher quality and more stable, our human evaluation focused on GPT-NIP data to confirm the quality of the best synthetic data. We invited two medical students with clinical experience and three undergraduate students to conduct evaluations, with an equal split of 42 instances for both Q\&A and explanation tasks (totally 84 cases). Among the 66 instances evaluated on GPT-4 NIP data, 52 instances had EHR notes from discharge instructions, while 14 were from MADE . The human evaluation results are presented in \tableref{tab:evalOverview}.

\section{Limitation}
Our pipeline was only validated on the  MIMIC-III and MADE 1.0 datasets, and its performance may not necessarily generalize to different types of datasets and medical environments. Despite the good performance of LLMs in the two tasks, hallucination issues still persist in interactions. Additionally, the explanations generated by LLMs cannot guarantee that patients will correctly understand them, requiring careful consideration in practical deployment. Besides, the amount of human evaluation is relatively insufficient at present. In the subsquential plan, more human evaluation can be added. The limitation mentioned above will be the focus of our upcoming work, which includes training our own in-house system.

\section{Conclusion}
Taking into account the results from both LLM evaluation and human evaluation, it is evident that the NoteAid EHR Interaction Dataset generated by GPT-4 NIP, whether in the Q\&A task or the Explanation task, consistently exhibits high quality. The proportion of data meeting the highest quality level exceeds 80\%. This demonstrates that our NoteAid EHR Interaction Pipeline performs well in both tasks, affirming the capability of LLMs to provide interactive services with patients regarding their medical notes and enhance the level of patient medical education.

\bibliography{jmlr-sample}

\newpage
\appendix

\section{Creteria Overview}
\begin{figure}[h]
    \centering
    \floatconts
    {fig:qualityLevel}
    {\caption{\textbf{Evaluation Quality Level}}}
    {\includegraphics[width=0.6\textwidth]{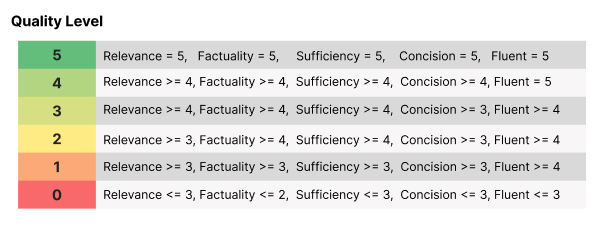}}
\end{figure}
\begin{table}[h]
\floatconts
    {tab:criteria}
    {\caption{\textbf{Criteria for the evaluation} Both human evaluation and llm evaluation mainly followed this criteria.}}
    {
    \begin{tabular}{p{14.6cm}}
    \hline
    \textbf{Criteria Overview}\\
    \hline
    \hline
    \textbf{Relevance}\\
     1. An answer that fully focus on the question, without off topic result worth 5 points. Eg. A question about ERCP may not have direct relevance to vitamin B12. If an answer is explaining ERCP, it should not mention vitamin B12 if the patient didn't asked.\\
     2. Each irrelevant sentence results a deduction of 1 point.\\
    \hline
    \textbf{Factuality}\\
    1. Everything mentioned in the answer consistent with objective and correct medical knowledge worth 5 points.\\
    2. Each wrong medical knowledge in the answer results a deduction of at least 1 point (Score according to the impact of the error, A failure that won't influence the patient to understand his or her own illness result a deduction of 1 point, while errors that may mislead are penalized more than 1 point based on their severity).\\
    \hline
    \textbf{Sufficiency}\\
    1. An sufficient answer should cover all patient's confusion mentioned in the question. All points has been answered worth 5 points.  \\
    2. Each missed point result a deduction of 1 point. \\
    \hline
    \textbf{Concision}\\ 
    1. A concise and clear syntax and vocabulary, devoid of unnecessary conversation and filler words like ``I'm happy to help,'' worth 5 points.  \\
    2. Each redundant sentence in the answer results a deduction of 1 point.\\
    \hline
    \textbf{Fluent}\\
    Is the language fluent and easy to understand? Nothing vague or hard to understand worth 5 points. Scoring according to the actual situation of your own reading process. \\
    \hline
    \end{tabular}
    }
\end{table}

\newpage
\section{Prompt Overview}
\begin{table}[h]
    \centering
    \begin{tabular}{p{14.6cm}}
    \toprule
    \textbf{Mock Patient Agent Generation Prompt Overview}\\
    \midrule
    \midrule
    \textbf{System Prompt for both Mock Patient Agent and Assistant Agent}\\
    Reference Content Including\\
    Medical Notes:\\
    $<$EHR Note content$>$\\
    \midrule
    \textbf{Mock Patient Agent initial user prompt for Q\&A task}\\
    Try to mock as the a patient and ask one question that the patient may not understand.\\
    Return the output as a dictionary object, adhering to the following structure:\\
    \{``question'': $<$mock question content that patient may ask$>$\}\\
    Provide your response solely in the dictionary without any additional text.\\
    \midrule
    \textbf{Mock Patient Agent follow up user prompt for Q\&A task}\\
    Here is the answer for the question that mentioned above:\\
    $<$Answer provided by the Assistant Agent will be listed here$>$\\
    Try to mock as the a patient and ask a new question that the patient may not understand.\\
    Return the output as a dictionary object with the same format above.\\
    \midrule
    \textbf{Mock Patient Agent initial user prompt for explanation task}\\
    Try to mock as the a patient and select one sentence from the medical note that the patient may not understand.\\
    Return the output as a dictionary object, adhering to the following structure:\\
    \{``content'': $<$origin content that patient may not understand$>$\}\\
    Provide your response solely in the dictionary without any additional text.\\
    \midrule
    \textbf{Mock Patient Agent follow up user prompt for explanation task}\\
    Here is the explanation for the content that mentioned above:\\
    $<$Explanation provided by the Assistant Agent will be listed here$>$\\
    Try to mock as the a patient and select a new sentence from the medical note that the patient may not understand.\\
    Return the output as a dictionary object with the same format above.\\
    \bottomrule
    \end{tabular}
    \caption{\textbf{Mock Patient Agent Generation Prompt} Since we are using gpt-3.5-turbo and gpt-4, they both accept system prompt and user prompt. So we designed the same prompt.}
    \label{tab:generationPrompt}
\end{table}

\begin{table}[h]
    \centering
    \begin{tabular}{p{14.6cm}}
    \toprule
    \textbf{Assistant Agent Generation Prompt Overview}\\
    \midrule
    \midrule
    \textbf{Assistant Agent initial user prompt for Q\&A task}\\
    Here is the question:\\
    $<$Question provided by Mock Patient Agent will be listed here$>$\\
    Answer the question based on the reference content and use concise language that people are easy to understand.\\
    Your answers should be very careful to ensure that the questions asked by the patient do not conflict with the medical note. Mark answers you are not sure about.\\
    \midrule
    \textbf{Assistant Agent follow up user prompt for Q\&A task}\\
    Here is another question:\\
    $<$Question provided by Mock Patient Agent will be listed here$>$\\
    Answer the question based on the reference content and use concise language that people are easy to understand.\\
    Your answers should be very careful to ensure that the questions asked by the patient do not conflict with the medical note. Mark answers you are not sure about.\\
    \midrule
    \textbf{Assistant Agent initial user prompt for explanation task}\\
    Here is the origin content from the medical note:\\
    $<$Content provided by Mock Patient Agent will be listed here$>$\\
    Explain the content for the patient based on the reference content and use concise language that people are easy to understand.\\
    Your answers should be very careful to ensure that the questions asked by the patient do not conflict with the medical note. Mark answers you are not sure about.\\
    \midrule
    \textbf{Assistant Agent follow up user prompt for explanation task}\\
    Here is annother origin content from the medical note:\\
    $<$Content provided by Mock Patient Agent will be listed here$>$\\
    Explain the content for the patient based on the reference content and use concise language that people are easy to understand.\\
    Your answers should be very careful to ensure that the questions asked by the patient do not conflict with the medical note. Mark answers you are not sure about.\\
    \bottomrule
    \end{tabular}
    \caption{\textbf{Assistant Agent Generation Prompt}}
\end{table}

\begin{table}[h]
    \label{tab:evalPrompt}
    \centering
    \begin{tabular}{p{14.6cm}}
    \toprule
    \textbf{Evaluation Prompt}\\
    \midrule
    \midrule
    \textbf{System Prompt Overview}\\
    We have Q\&A conversations and content Explanations based on the given medical note. \\
    A Q\&A conversation includes a question asked by the patient and an answer answered by the assistant. 
    An explanation includes a selected content represent part of medical note that hard to be understood by the patient and its explanation provided by the assistant.\\
    Give a feedback of the performance of assistant in each conversation or explanation follow the criteria:\\
    \\
    Relevance:\\
    1. An answer that fully focus on the question, without off topic result worth 5 points. Eg. A question about ERCP may not have direct relevance to vitamin B12. If an answer is explaining ERCP, it should not mention vitamin B12 if the patient didn't asked.  \\
    2. Each irrelevant sentence results a deduction of 1 point.  \\
    Factuality: \\
    1. Everything mentioned in the answer consistent with objective and correct medical knowledge worth 5 points.\\
    2. Each wrong medical knowledge in the answer results a deduction of at least 1 point (Score according to the impact of the error, A failure that won't influence the patient to understand his or her own illness result a deduction of 1 point, while errors that may mislead are penalized more than 1 point based on their severity).  \\
    Sufficiency:  \\
    1. An sufficient answer should cover all patient's confusion mentioned in the question. All points has been answered with logic worth 5 points.  \\
    2. Each missed point result a deduction of 1 point. \\
    Concision: \\
    1. A concise and clear syntax and vocabulary, devoid of unnecessary conversation and filler words like ``I'm happy to help,'' worth 5 points.  \\
    2. Each redundant sentence in the answer results a deduction of 1 point.\\
    Fluent: \\
    Is the language fluent and easy to understand? Nothing vague or hard to understand worth 5 points. Scoring according to the actual situation of your own reading process.  \\
    \\
    Try not give full credits, full credits means perfect. If you found any incompleteness, make a deduction.\\
    \\
    Generate the result in a dictionary format as the following example:\\
    {``Relevance'': 4, ``Factuality'': 5, ``Sufficiency'': 4, ``Concision'': 3, ``Fluent'': 5}\\
    \\
    Provide your response solely in the dictionary without any additional text.\\
    \bottomrule
    \end{tabular}
    \caption{\textbf{Evaluation Prompt Part 1} The prompt that applied in llm evaluation. Mainly described criteria shows above with some instructions}
\end{table}

\begin{table}[h]
    \centering
    \begin{tabular}{p{14.6cm}}
    \toprule
    \textbf{User Prompt 1}\\
    Here is the medical note:\\
    $<$EHR Note content$>$\\
    Here is the first conversation (explanation), try to be strict:\\
    $<$The first question with answer or content with explanation will be listed here$>$\\
    \midrule
    \textbf{User Prompt 2}\\
    Here is another conversation (explanation) based on the given medical note, try to be strict:\\
    $<$Second and third conversation or explanation will be listed here with the same prompt$>$\\
    \bottomrule
    \end{tabular}
    \label{tab:evalPrompt2}
    \caption{\textbf{Evaluation Prompt Part 2} The prompt that applied in llm evaluation. Mainly described criteria shows above with some instructions}
\end{table}

\newpage
\section{Case Study}
\begin{table}[h]
    \centering
    \begin{tabular}{p{14.6cm}}
    \toprule
    \textbf{Explanation Example}\\
    \midrule
    \midrule
    \textbf{EHR Note}\\
    ...Your INR was found to be high so your coumadin was stopped and you were transfused blood and clotting factors...\\
    \textbf{Mock Patient Agent selected sontent}\\
    Your INR was found to be high so your coumadin was stopped and you were transfused blood and clotting factors.\\
    \textbf{Assistant Agent's explanation}\\
    During your hospital stay, we discovered that a blood test called INR, which measures how quickly your blood clots, was higher than normal. This could increase your risk of bleeding. As a result, we stopped your medication called coumadin, which is a blood thinner, and gave you a blood transfusion and clotting factors to help your blood clot more normally.\\
    \midrule
    \textbf{EHR Note}\\ 
    ...The patient was treated with a course of steroids and due to the fall of the patient's hemoglobin and hematocrit, the patient was transferred to [** Hospital **] and transferred to the medical wards...\\
    \textbf{Mock Patient Agent selected content}\\
    The patient was treated with a course of steroids and due to the fall of the patient's hemoglobin and hematocrit, the patient was transferred to [** Hospital **] and transferred to the medical wards.\\
    \textbf{Assistant Agent's explanation}\\
    You were given a course of medication known as steroids, which are often used to reduce inflammation and suppress the immune system. \textcolor{red}{This was because your hemoglobin and hematocrit levels had fallen}. Hemoglobin is a protein in your red blood cells that carries oxygen, and hematocrit is a measure of how much of your blood is made up of red blood cells. When these levels fall, it can indicate a problem like anemia or blood loss. Due to these issues, you were moved to another hospital and admitted to their medical wards for further care and treatment.\\
    \bottomrule
    \end{tabular}
    \caption{\textbf{Case Study 1} Here are two explanation examples. The first example serves as a positive instance, while the second one functions as a negative instance. The red highlight sentence in the explanation is irrelevant to this selected content.``Reduce inflammation and suppress the immune system.'' is not a result of ``hemoglobin and hematocrit levels had fallen''. This is a medical factual error.}
\end{table}

\clearpage
\begin{table}[h]
    \centering
    \begin{tabular}{p{14.6cm}}
    \toprule
    \textbf{Q\&A Example}\\
    \midrule
    \midrule
    \textbf{EHR Note}\\
    ...Since you had an allergic reaction to either bactrim (TMP-SMX) or keflex (cephalexin), we recommend that you DO NOT take these medications again in the future...\\
    \textbf{Mock Patient Agent question}\\
    Why did I develop an allergic reaction to the antibiotics? I've taken them before without any problems.\\
    \textbf{Assistant Agent's answer}\\
    Allergic reactions can develop even if you have taken a medication before without any problems. Sometimes, the body's immune system can become sensitized to a medication over time, leading to an allergic reaction upon subsequent exposure. It is also possible for allergic reactions to develop suddenly, even if you have never had a reaction to a medication before. It is important to inform your healthcare provider about any allergies or adverse reactions you experience so they can make appropriate adjustments to your treatment plan.\\
    \midrule
    \textbf{EHR Note}\\
    ...The opthalmologists here prescribed you eye drops to take when you go home, and they ask that you follow up with them in their clinic next week...\\
    \textbf{Mock Patient Agent question}\\ 
    Why do I need to follow up with the ophthalmologist for my eye redness? Can't I just use the prescribed eye drops at home?\\
    \textbf{Assistant Agent's answer}\\
    It is important to follow up with the ophthalmologist for your eye redness because they are specialists in eye health and can provide a thorough evaluation of your condition. While the prescribed eye drops may help alleviate the symptoms, the ophthalmologist will assess the effectiveness of the treatment and ensure that there are no underlying issues or complications that need to be addressed. They will also determine if any further treatment or adjustments to the medication are necessary.\\
    \bottomrule
    \end{tabular}
    \caption{\textbf{Case Study 2} Here are also two explanation examples. The first example serves as a positive instance, while the second one functions as a negative instance. The answer here is not combined the case to explain enough}
\end{table}

\end{document}